# Operations on soft sets revisited

Ping Zhu[a,b,*], Qiaoyan Wen[b]

[a]*School of Science, Beijing University of Posts and Telecommunications, Beijing 100876, China*
[b]*State Key Laboratory of Networking and Switching, Beijing University of Posts and Telecommunications, Beijing 100876, China*

**Abstract**

Soft sets, as a mathematical tool for dealing with uncertainty, have recently gained considerable attention, including some successful applications in information processing, decision, demand analysis, and forecasting. To construct new soft sets from given soft sets, some operations on soft sets have been proposed. Unfortunately, such operations cannot keep all classical set-theoretic laws true for soft sets. In this paper, we redefine the intersection, complement, and difference of soft sets and investigate the algebraic properties of these operations along with a known union operation. We find that the new operation system on soft sets inherits all basic properties of operations on classical sets, which justifies our definitions.

## 1. Introduction

As a necessary supplement to some existing mathematical tools for handling uncertainty, Molodtsov [17] initiated the concept of soft sets via a set-valued mapping. A distinguishing feature of soft sets which is different from probability theory, fuzzy sets, and interval mathematics is that precise quantity such as probability and membership grade is not essential. This feature facilitates some applications because in most realistic settings the underlying probabilities and membership grades are not known with sufficient precision to justify the use of numerical valuations. Since its introduction, the concept of soft sets has gained considerable attention (see, for example, [1, 4, 8, 9, 10, 11, 12, 13, 14, 16, 20, 22, 24, 25]), including some successful applications in information processing [6, 7, 18, 26], decision [3, 15, 19], demand analysis [5], and forecasting [21].

In [14], Maji, Biswas, and Roy made a theoretical study of the soft set theory in more detail. Especially, they introduced the concepts of subset, intersection, union, and complement of soft sets and discussed their properties. These operations make it possible to construct new soft sets from given soft sets. Unfortunately, several basic properties presented in [14] are not true in general; these have been pointed out and improved by Yang [23] and Ali et al. [2]. In particular, to keep some classical set-theoretic laws true for soft sets, Ali et al. defined some restricted operations on soft sets such as the restricted intersection, the restricted union, and the restricted difference and improved the notion of complement of a soft set. Based upon these newly defined operations, they proved that certain De Morgan's laws hold for soft sets. It is worth noting that the concept of complement [2, 14] [1] which is fundamental to De Morgan's laws is based on the so-called NOT set of a parameter set. It means that the logic conjunction NOT is a prerequisite for defining the complement of a soft set; this is considerably beyond the definition of soft sets. Moreover, the union of a soft set and its complement is not exactly the whole universal soft set in general, which is considered less desirable.

The purpose of this paper is to develop the theory of soft sets by introducing new operations on soft sets that inherit all basic properties of classical set operations. To this end, we redefine the intersection, complement, and difference of soft sets and then examine the algebraic properties of these operations along with a known union operation. It turns out that all basic properties of operations on classical sets, including identity laws, domination laws, idempotent laws, commutative laws, associative laws, distributive laws, and De Morgan's laws, hold for soft sets with respect to the newly defined operations.

---

*Corresponding author
*Email addresses:* `pzhubupt@gmail.com` (Ping Zhu), `wqy@bupt.edu.cn` (Qiaoyan Wen)

[1]It should be stressed that there are two types of complements defined in [2]: one is defined with the NOT set of parameters and the other is defined without the NOT set of parameters.

The remainder of this paper is structured as follows. In Section 2, we briefly recall the notion of soft sets. Section 3 is devoted to the definitions of new operations on soft sets. An example is also provided to illustrate the newly defined operations in this section. We address the basic properties of the operations on soft sets in Section 4 and conclude the paper in Section 5.

## 2. Soft sets

For subsequent need, let us review the notion of soft sets. For a detailed introduction to the soft set theory, the reader may refer to [14, 17].

We begin with some notations. For classical set theory, the symbols $\emptyset$, $A^c$, $A \cup B$, $A \cap B$, $A \setminus B$ denote, respectively, the empty set, the complement of $A$ with respect to some universal set $U$, the union of sets $A$ and $B$, the intersection of sets $A$ and $B$, and the difference of sets $A$ and $B$ whose elements belong to $A$ but not to $B$. In what follows, we write $\mathscr{P}(U)$ for the power set of a universal set $U$ and denote $\mathscr{P}(U) \setminus \{\emptyset\}$ by $\mathscr{P}^*(U)$.

Throughout this paper, let $U$ be a universal set and $E$ be the set of all possible parameters under consideration with respect to $U$. Usually, parameters are attributes, characteristics, or properties of objects in $U$. We now recall the notion of soft sets due to Molodtsov [17].

**Definition 2.1** (Molodtsov [17]). Let $U$ be a universe and $E$ the set of parameters. A *soft set* over $U$ is a pair $(F, A)$ consisting of a subset $A$ of $E$ and a mapping $F : A \longrightarrow \mathscr{P}^*(U)$.

Note that the above definition is slightly different from the original one in [17] where $F$ has $\mathscr{P}(U)$ as its codomain. In other words, we remove the parameters having the empty set as images under $F$. It seems rational since this means that if there exists a parameter $e \in A$ which is not the attribute, characteristic, or property of any object in $U$, then this parameter has no interest with respect to the knowledge stored in the soft set. As a result, a soft set of $U$ in the sense of Definition 2.1 is a parameterized family of nonempty subsets of $U$.

Clearly, any soft set $(F, A)$ over $U$ gives a partial function $F' : E \longrightarrow \mathscr{P}^*(U)$ defined by

$$F'(e) = \begin{cases} F(e), & \text{if } e \in A \\ \text{undefined}, & \text{otherwise} \end{cases}$$

for all $e \in E$. Conversely, any partial function $f$ from $E$ to $\mathscr{P}^*(U)$ gives rise to a soft set $(F_f, A_f)$, where $A_f = \{e \in E \mid f(e) \text{ is defined}\}$ and $F_f$ is the restriction of $f$ on $A_f$.

To illustrate the above definition, Molodtsov considered several examples in [17], one of which we present blow.

**Example 2.2.** Suppose that $U$ is the set of houses under consideration, say $U = \{h_1, h_2, \ldots, h_5\}$. Let $E$ be the set of some attributes of such houses, say $E = \{e_1, e_2, \ldots, e_8\}$, where $e_1, e_2, \ldots, e_8$ stand for the attributes "expensive", "beautiful", "wooden", "cheap", "in the green surroundings", "modern", "in good repair", and "in bad repair", respectively.

In this case, to define a soft set means to point out expensive houses, beautiful houses, and so on. For example, the soft set $(F, A)$ that describes the "attractiveness of the houses" in the opinion of a buyer, say Alice, may be defined like this:

$$A = \{e_2, e_3, e_4, e_5, e_7\};$$
$$F(e_2) = \{h_2, h_3, h_5\}, \ F(e_3) = \{h_2, h_4\}, \ F(e_4) = \{h_1\}, \ F(e_5) = U, \ F(e_7) = \{h_3, h_5\}.$$

## 3. Operations on soft sets

In this section, we generalize the basic operations on classical sets to soft sets. The examination of more properties of these operations is deferred to the next section.

Let us start with the notions of empty and universal soft sets. Recall that in [14] a soft set $(F, A)$ is called a null soft set if $F(e) = \emptyset$ for all $e \in A$. Because $\emptyset$ does not belong to the codomain of $F$ in our framework, we redefine the concept of empty soft set as follows.

**Definition 3.1.** A soft set $(F, A)$ over $U$ is said to be *empty* whenever $A = \emptyset$. Symbolically, we write $(\emptyset, \emptyset)$ for the empty soft set over $U$.

**Definition 3.2.** A soft set $(F, A)$ over $U$ is called a *universal soft set* if $A = E$ and $F(e) = U$ for all $e \in A$. Symbolically, we write $(U, E)$ for the universal soft set over $U$.

Let us now define the subsets of a soft set.

**Definition 3.3.** Let $(F, A)$ and $(G, B)$ be two soft sets over $U$. We say that $(F, A)$ is a *subset* of $(G, B)$, denoted $(F, A) \subseteq (G, B)$, if either $(F, A) = (\emptyset, \emptyset)$ or $A \subseteq B$ and $F(e) \subseteq G(e)$ for every $e \in A$. Two soft sets $(F, A)$ and $(G, B)$ are said to be *equal*, denoted $(F, A) = (G, B)$, if $(F, A) \subseteq (G, B)$ and $(G, B) \subseteq (F, A)$.

By definition, two soft sets $(F, A)$ and $(G, B)$ are equal if and only if $A = B$ and $F(e) = G(e)$ for all $e \in A$. In [14], a similar notion, called soft subset, was defined by requiring that $A \subseteq B$ and $F(e) = G(e)$ for every $e \in A$. By Definition 3.3, the empty soft set $(\emptyset, \emptyset)$ is a subset of any soft set. It also follows from Definition 3.3 that any soft set is a subset of the universal soft set $(U, E)$. Formally, we have the following proposition.

**Proposition 3.4.** *For any soft set $(F, A)$ over $U$,*

$$(\emptyset, \emptyset) \subseteq (F, A) \subseteq (U, E).$$

We are now in the position to introduce some operations on soft sets.

**Definition 3.5.** Let $(F, A)$ and $(G, B)$ be two soft sets over $U$. The *intersection* of $(F, A)$ and $(G, B)$, denoted by $(F, A) \cap (G, B)$, is defined as $(F \cap G, C)$, where

$$C = \{e \in A \cap B \mid F(e) \cap G(e) \neq \emptyset\}, \text{ and}$$
$$(F \cap G)(e) = F(e) \cap G(e), \text{ for all } e \in C.$$

In particular, if $A \cap B = \emptyset$ or $F(e) \cap G(e) = \emptyset$ for every $e \in A \cap B$, then we see that $(F, A) \cap (G, B) = (\emptyset, \emptyset)$.

The following definition of union of soft sets is the same as in [14].

**Definition 3.6** ([14], Definition 2.11)**.** Let $(F, A)$ and $(G, B)$ be two soft sets over $U$. The *union* of $(F, A)$ and $(G, B)$, denoted by $(F, A) \cup (G, B)$, is defined as $(F \cup G, C)$, where

$$C = A \cup B, \text{ and for all } e \in C,$$
$$(F \cup G)(e) = \begin{cases} F(e), & \text{if } e \in A \setminus B \\ G(e), & \text{if } e \in B \setminus A \\ F(e) \cup G(e), & \text{otherwise.} \end{cases}$$

We now define the notion of complement in soft set theory. It is worth noting that this is rather different from those in the existing literature [14, 2], where the complement is usually based on the so-called NOT set of a parameter set and the union of a soft set and its complement is not exactly the whole universal soft set in general.

**Definition 3.7.** Let $(F, A)$ be a soft set over $U$. The *complement* of $(F, A)$ with respect to the universal soft set $(U, E)$, denoted by $(F, A)^c$, is defined as $(F^c, C)$, where

$$C = E \setminus \{e \in A \mid F(e) = U\} = \{e \in A \mid F(e) = U\}^c, \text{ and for all } e \in C,$$
$$F^c(e) = \begin{cases} U \setminus F(e), & \text{if } e \in A \\ U, & \text{otherwise.} \end{cases}$$

In certain settings, the difference of two soft sets $(F, A)$ and $(G, B)$ is useful.

**Definition 3.8.** Let $(F, A)$ and $(G, B)$ be two soft sets over $U$. The *difference* of $(F, A)$ and $(G, B)$, denoted by $(F, A) \setminus (G, B)$, is defined as $(F \setminus G, C)$, where

$$C = A \setminus \{e \in A \cap B \mid F(e) \subseteq G(e)\}, \text{ and for all } e \in C,$$
$$(F \setminus G)(e) = \begin{cases} F(e) \setminus G(e), & \text{if } e \in A \cap B \\ F(e), & \text{otherwise.} \end{cases}$$

By Definitions 3.7 and 3.8, we find that $(F, A)^c = (U, E)\setminus(F, A)$ holds for any soft set $(F, A)$. That is, the complement of $(F, A)$ with respect to the universal soft set $(U, E)$ is exactly the difference of $(U, E)$ and $(F, A)$. In light of this, $(F, A)\setminus(G, B)$ is also called the relative complement of $(G, B)$ in $(F, A)$), while $(F, A)^c$ is also called the absolute complement of $(F, A)$.

Let us illustrate the previous operations on soft sets by a simple example.

**Example 3.9.** Let us revisit Example 2.2. Recall that the soft set $(F, A)$ describing the "attractiveness of the houses" in Alice's opinion was defined by

$$A = \{e_2, e_3, e_4, e_5, e_7\};$$
$$F(e_2) = \{h_2, h_3, h_5\},\ F(e_3) = \{h_2, h_4\},\ F(e_4) = \{h_1\},\ F(e_5) = U,\ F(e_7) = \{h_3, h_5\}.$$

In addition, we assume that the "attractiveness of the houses" in the opinion of another buyer, say Bob, is described by the soft set $(G, B)$, where

$$B = \{e_1, e_2, \ldots, e_7\};$$
$$G(e_1) = \{h_3, h_5\},\ G(e_2) = \{h_4\},\ G(e_3) = \{h_2, h_4\},\ G(e_4) = \{h_1\},\ G(e_5) = \{h_2, h_3, h_4, h_5\},\ G(e_6) = G(e_7) = \{h_3\}.$$

Then by a direct computation, we can readily obtain $(F, A) \cap (G, B)$, $(F, A) \cup (G, B)$, $(F, A)^c$, and $(F, A)\setminus(G, B)$ as follows:

- $(F, A) \cap (G, B) = (F \cap G, \{e_3, e_4, e_5, e_7\})$, where $(F \cap G)(e_3) = \{h_2, h_4\}$, $(F \cap G)(e_4) = \{h_1\}$, $(F \cap G)(e_5) = \{h_2, h_3, h_4, h_5\}$, and $(F \cap G)(e_7) = \{h_3\}$. This means that both Alice and Bob think that $h_2$ and $h_4$ are wooden, $h_1$ is cheap, $h_2, h_3, h_4, h_5$ are in the green surroundings, and $h_3$ is in the good repair.

- $(F, A) \cup (G, B) = (F \cup G, \{e_1, e_2, \ldots, e_7\})$, where $(F \cup G)(e_1) = \{h_3, h_5\}$, $(F \cup G)(e_2) = \{h_2, h_3, h_4, h_5\}$, $(F \cup G)(e_3) = \{h_2, h_4\}$, $(F \cup G)(e_4) = \{h_1\}$, $(F \cup G)(e_5) = U$, $(F \cup G)(e_6) = \{h_3\}$, and $(F \cup G)(e_7) = \{h_3, h_5\}$. This means that either Alice or Bob thinks that $h_3$ is expensive, either Alice or Bob thinks that $h_5$ is expensive, either Alice or Bob thinks that $h_2$ is beautiful, either Alice or Bob thinks that $h_3$ is beautiful, and so on.

- $(F, A)^c = (F^c, \{e_1, e_2, e_3, e_4, e_6, e_7, e_8\})$, where $F^c(e_1) = U$, $F^c(e_2) = \{h_1, h_4\}$, $F^c(e_3) = \{h_1, h_3, h_5\}$, $F^c(e_4) = \{h_2, h_3, h_4, h_5\}$, $F^c(e_6) = U$, $F^c(e_7) = \{h_1, h_2, h_4\}$, and $F^c(e_8) = U$. This means that Alice thinks that none of these houses is expensive, neither $h_1$ nor $h_4$ is beautiful, $h_1, h_3, h_5$ are not wooden, and so on.

- $(F, A)\setminus(G, B) = (F\setminus G, \{e_2, e_5, e_7\})$, where $(F\setminus G)(e_2) = \{h_2\}$, $(F\setminus G)(e_5) = \{h_1\}$, and $(F\setminus G)(e_7) = \{h_5\}$. This means that Alice thinks of $h_2$ as beautiful, but Bob does not think that $h_2$ is beautiful, and so on.

## 4. Algebraic properties of soft set operations

This section is devoted to some algebraic properties of soft set operations defined in the last section.

Let us begin with some properties involving intersections and unions. The first four laws are obvious. We omit their proofs here since the proofs follow directly from the definitions of intersection and union of soft sets.

**Proposition 4.1** (Identity laws). *For any soft set $(F, A)$ over $U$, we have that*

(1) $(F, A) \cap (U, E) = (F, A)$.
(2) $(F, A) \cup (\emptyset, \emptyset) = (F, A)$.

**Proposition 4.2** (Domination laws). *For any soft set $(F, A)$ over $U$, we have that*

(1) $(F, A) \cap (\emptyset, \emptyset) = (\emptyset, \emptyset)$.
(2) $(F, A) \cup (U, E) = (U, E)$.

**Proposition 4.3** (Idempotent laws). *For any soft set $(F, A)$ over $U$, we have that*

(1) $(F, A) \cap (F, A) = (F, A)$.

(2) $(F, A) \cup (F, A) = (F, A)$.

**Proposition 4.4** (Commutative laws). *For any soft sets $(F, A)$ and $(G, B)$ over $U$, we have that*

(1) $(F, A) \cap (G, B) = (G, B) \cap (F, A)$.
(2) $(F, A) \cup (G, B) = (G, B) \cup (F, A)$.

Now we turn our attention to the associative laws.

**Proposition 4.5** (Associative laws). *For any soft sets $(F, A)$, $(G, B)$, and $(H, C)$ over $U$, we have that*

(1) $((F, A) \cap (G, B)) \cap (H, C) = (F, A) \cap ((G, B) \cap (H, C))$.
(2) $((F, A) \cup (G, B)) \cup (H, C) = (F, A) \cup ((G, B) \cup (H, C))$.

*Proof.* We only prove the first assertion, since the second one is the same as Proposition 2.5(i) in [14]. For simplicity, we write $(L, A')$, $(R, B')$, and $(F \cap G, A_1)$ for $((F, A) \cap (G, B)) \cap (H, C)$, $(F, A) \cap ((G, B) \cap (H, C))$, and $(F, A) \cap (G, B)$, respectively. We thus get by definition that

$$
\begin{aligned}
A' &= \{e \in A_1 \cap C \mid (F \cap G)(e) \cap H(e) \neq \emptyset\} \\
&= \{e \in A_1 \mid (F \cap G)(e) \cap H(e) \neq \emptyset\} \cap \{e \in C \mid (F \cap G)(e) \cap H(e) \neq \emptyset\} \\
&= \{e \in A \cap B \mid (F \cap G)(e) \neq \emptyset, (F \cap G)(e) \cap H(e) \neq \emptyset\} \cap \{e \in C \mid (F \cap G)(e) \cap H(e) \neq \emptyset\} \\
&= \{e \in A \cap B \mid (F \cap G)(e) \cap H(e) \neq \emptyset\} \cap \{e \in C \mid (F \cap G)(e) \cap H(e) \neq \emptyset\} \\
&= \{e \in A \cap B \cap C \mid (F \cap G)(e) \cap H(e) \neq \emptyset\} \\
&= \{e \in A \cap B \cap C \mid F(e) \cap G(e) \cap H(e) \neq \emptyset\}.
\end{aligned}
$$

By the same token, we have that $B' = \{e \in A \cap B \cap C \mid F(e) \cap G(e) \cap H(e) \neq \emptyset\}$, and thus $A' = B'$. Moreover, for any $e \in A'$, we have that

$$
\begin{aligned}
L(e) &= (F \cap G)(e) \cap H(e) \\
&= F(e) \cap G(e) \cap H(e) \\
&= F(e) \cap (G(e) \cap H(e)) \\
&= F(e) \cap (G \cap H)(e) \\
&= R(e),
\end{aligned}
$$

namely, $L(e) = R(e)$. Therefore, the assertion (1) holds. □

**Proposition 4.6** (Distributive laws). *For any soft sets $(F, A)$, $(G, B)$, and $(H, C)$ over $U$, we have that*

(1) $(F, A) \cap ((G, B) \cup (H, C)) = ((F, A) \cap (G, B)) \cup ((F, A) \cap (H, C))$.
(2) $(F, A) \cup ((G, B) \cap (H, C)) = ((F, A) \cup (G, B)) \cap ((F, A) \cup (H, C))$.

*Proof.* We only verify the first assertion; the second one can be verified similarly. For simplicity, we write $(L, A')$ and $(R, B')$ for $(F, A) \cap ((G, B) \cup (H, C))$ and $((F, A) \cap (G, B)) \cup ((F, A) \cap (H, C))$, respectively. We thus see that

$$
\begin{aligned}
A' &= \{e \in A \cap (B \cup C) \mid F(e) \cap (G \cup H)(e) \neq \emptyset\} \\
&= \{e \in (A \cap B) \cup (A \cap C) \mid F(e) \cap (G \cup H)(e) \neq \emptyset\} \\
&= \{e \in A \cap B \mid F(e) \cap (G \cup H)(e) \neq \emptyset\} \cup \{e \in A \cap C \mid F(e) \cap (G \cup H)(e) \neq \emptyset\} \\
&= \{e \in A \cap B \cap C^c \mid F(e) \cap G(e) \neq \emptyset\} \cup \{e \in A \cap B \cap C \mid F(e) \cap (G(e) \cup H(e)) \neq \emptyset\} \\
&\quad \cup \{e \in A \cap B^c \cap C \mid F(e) \cap H(e) \neq \emptyset\} \\
&= \{e \in A \cap B \cap C^c \mid F(e) \cap G(e) \neq \emptyset\} \cup \{e \in A \cap B \cap C \mid F(e) \cap G(e) \neq \emptyset\} \\
&\quad \cup \{e \in A \cap B^c \cap C \mid F(e) \cap H(e) \neq \emptyset\} \cup \{e \in A \cap B \cap C \mid F(e) \cap H(e) \neq \emptyset\} \\
&= \{e \in A \cap B \mid F(e) \cap G(e) \neq \emptyset\} \cup \{e \in A \cap C \mid F(e) \cap H(e) \neq \emptyset\} \\
&= B',
\end{aligned}
$$

namely, $A' = B'$. Furthermore, for any $e \in A'$, one can check that $L(e) = F(e) \cap (G \cup H)(e) = ((F \cap G) \cup (F \cap H))(e) = R(e)$ by a routine computation. We do not go into the details here. Hence, the assertion (1) holds. □

Like usual sets, soft sets are monotonic with respect to intersection and union.

**Proposition 4.7.** *Let $(F_i, A_i)$ and $(G_i, B_i)$, $i = 1, 2$, be soft sets over $U$. If $(F_i, A_i) \subseteq (G_i, B_i)$, $i = 1, 2$, then we have that $(F_1, A_1) \cap (F_2, A_2) \subseteq (G_1, B_1) \cap (G_2, B_2)$ and $(F_1, A_1) \cup (F_2, A_2) \subseteq (G_1, B_1) \cup (G_2, B_2)$.*

*Proof.* It is clear by the definitions of intersection, union, and subset of soft sets. □

Recall that in classical set theory, we have that $X \subseteq Y$ if and only if $X \cap Y = X$, which is also equivalent to $X \cup Y = Y$. For soft sets, we have the following observation.

**Proposition 4.8.** *Let $(F, A)$ and $(G, B)$ be soft sets over $U$. Then the following are equivalent.*

(1) $(F, A) \subseteq (G, B)$.
(2) $(F, A) \cap (G, B) = (F, A)$.
(3) $(F, A) \cup (G, B) = (G, B)$.

*Proof.* Again, it is obvious by the definitions of intersection, union, and subset of soft sets. □

The following several properties are concerned with the complement of soft sets.

**Proposition 4.9.** *Let $(F, A)$ and $(G, B)$ be two soft sets over $U$. Then $(G, B) = (F, A)^c$ if and only if $(F, A) \cap (G, B) = (\emptyset, \emptyset)$ and $(F, A) \cup (G, B) = (U, E)$.*

*Proof.* If $(G, B) = (F, A)^c$, then we see by definition that $(F, A) \cap (F, A)^c = (\emptyset, \emptyset)$ and $(F, A) \cup (F, A)^c = (F, A) \cup (F^c, \{e \in A \mid F(e) = U\}^c) = (U, E)$. Whence, the necessity is true.

Conversely, assume that $(F, A) \cap (G, B) = (\emptyset, \emptyset)$ and $(F, A) \cup (G, B) = (U, E)$. The latter means that $A \cup B = E$. Moreover, we obtain that $F(e) = U$ for all $e \in A \backslash B$ and $G(e) = U$ for all $e \in B \backslash A$. For any $e \in A \cap B$, it follows from $(F, A) \cap (G, B) = (\emptyset, \emptyset)$ and $(F, A) \cup (G, B) = (U, E)$ that $F(e) \cup G(e) = U$ and $F(e) \cap G(e) = \emptyset$. As neither $F(e)$ nor $G(e)$ is empty, this forces that $B = \{e \in A \mid F(e) = U\}^c$. For any $e \in B$, if $e \in A$, then $G(e) = F(e)^c = F^c(e)$; if $e \in B \backslash A$, then $G(e) = U = F^c(e)$. This implies that $(F, A)^c = (F^c, \{e \in A \mid F(e) = U\}^c) = (G, B)$, finishing the proof. □

The following fact follows immediately from Proposition 4.9.

**Corollary 4.10.** *For any soft set $(F, A)$ over $U$, we have that $((F, A)^c)^c = (F, A)$.*

*Proof.* Note that $(F, A)^c \cap (F, A) = (\emptyset, \emptyset)$ and $(F, A)^c \cup (F, A) = (U, E)$. It therefore follows from Proposition 4.9 that $(F, A) = ((F, A)^c)^c$, as desired. □

With the above corollary, we can prove the De Morgan's laws of soft sets.

**Proposition 4.11** (De Morgan's laws). *For any soft sets $(F, A)$ and $(G, B)$ over $U$, we have that*

(1) $((F, A) \cap (G, B))^c = (F, A)^c \cup (G, B)^c$.
(2) $((F, A) \cup (G, B))^c = (F, A)^c \cap (G, B)^c$.

*Proof.* (1) For convenience, let $A_0 = \{e \in A \mid F(e) = U\}$, $B_0 = \{e \in B \mid G(e) = U\}$, $C_0 = \{e \in A \cap B \mid F(e) \cap G(e) = U\}$, and $C_1 = \{e \in A \cap B \mid F(e) \cap G(e) \neq \emptyset\}$. Then we have that

$$\begin{aligned} ((F, A) \cap (G, B))^c &= (F \cap G, C_1)^c \\ &= ((F \cap G)^c, \{e \in C_1 \mid (F \cap G)(e) = U\}^c) \\ &= ((F \cap G)^c, \{e \in A \cap B \mid F(e) \cap G(e) = U\}^c) \\ &= ((F \cap G)^c, C_0^c). \end{aligned}$$

On the other hand, we have that

$$\begin{aligned} (F, A)^c \cup (G, B)^c &= (F^c, A_0^c) \cup (G^c, B_0^c) \\ &= (F^c \cup G^c, A_0^c \cup B_0^c) \\ &= (F^c \cup G^c, (A_0 \cap B_0)^c) \\ &= (F^c \cup G^c, C_0^c). \end{aligned}$$

Therefore, to prove (1), it suffices to show that $(F\cap G)^c(e) = (F^c \cup G^c)(e)$ for all $e \in C_0^c$. In fact, since $C_0^c = (C_1\setminus C_0)\cup C_1^c$ and $(C_1\setminus C_0)\cap C_1^c = \emptyset$, we need only to consider two cases. The first case is that $e \in C_1\setminus C_0$. In this case, $e \in A_0^c \cap B_0^c$, and thus we get that $(F\cap G)^c(e) = (F(e)\cap G(e))^c = F(e)^c \cup G(e)^c = (F^c \cup G^c)(e)$. The other case is that $e \in C_1^c$. In this case, we always have by definition that $(F\cap G)^c(e) = U = (F^c \cup G^c)(e)$. Consequently, $(F\cap G)^c(e) = (F^c \cup G^c)(e)$ for all $e \in C_0^c$, as desired.

(2) By Corollary 4.10 and the first assertion, we find that

$$\begin{aligned}((F,A)\cup(G,B))^c &= (((F,A)^c)^c \cup ((G,B)^c)^c)^c \\ &= (((F,A)^c \cap (G,B)^c)^c)^c \\ &= (F,A)^c \cap (G,B)^c.\end{aligned}$$

Hence, the second assertion holds as well. This completes the proof of the proposition. □

Let us end this section with an observation on the difference of two soft sets.

**Proposition 4.12.** *For any soft sets $(F,A)$ and $(G,B)$ over U, we have that $(F,A)\setminus(G,B) = (F,A) \cap (G,B)^c$.*

*Proof.* We set $B_0 = \{e \in B \mid G(e) = U\}$ and write $(F\setminus G, C)$ for $(F,A)\setminus(G,B)$. Then we see that $C = A\setminus\{e \in A\cap B \mid F(e) \subseteq G(e)\}$ and $(G,B)^c = (G^c, B_0^c)$. As a result, $(F,A) \cap (G,B)^c = (F,A) \cap (G^c, B_0^c) = (F\cap G^c, B_1)$, where

$$\begin{aligned}B_1 &= \{e \in A\cap B_0^c \mid F(e)\cap G^c(e) \neq \emptyset\} \\ &= (A\setminus B) \cup \{e \in A\cap B \mid F(e) \nsubseteq G(e)\} \\ &= A\setminus\{e \in A\cap B \mid F(e) \subseteq G(e)\} \\ &= C,\end{aligned}$$

as desired. It remains to show that $(F\setminus G)(e) = (F\cap G^c)(e)$ for all $e \in C = B_1$. In fact, if $e \in C\setminus B$, then we have that $(F\setminus G)(e) = F(e) = F(e)\cap U = (F\cap G^c)(e)$; if $e \in C\cap B$, then $(F\setminus G)(e) = F(e)\setminus G(e) = F(e)\cap G^c(e) = (F\cap G^c)(e)$. We thus get that $(F\setminus G)(e) = (F\cap G^c)(e)$ for all $e \in C = B_1$. Consequently, $(F,A)\setminus(G,B) = (F,A)\cap(G,B)^c$, finishing the proof. □

## 5. Conclusion

In this paper, we have redefined the intersection, complement, and difference of soft sets. These operations, together with an existing union operation, form the fundamental operations for constructing new soft sets from given soft sets. By examining the algebraic properties of these operations, we find that all basic properties of operations on classical sets such as identity laws, domination laws, distributive laws, and De Morgan's laws hold for the newly defined operations on soft sets. From this point of view, the new operations on soft sets are reasonable. Motivated by the notion of Not set of a parameter set in [14], we will investigate the operations on soft sets by introducing more conjunctions including AND and OR into a parameter set. In addition, it is interesting to extend the notions of intersection, complement, difference of soft sets developed here to other soft structures such as fuzzy soft sets [19, 24], vague soft sets [22], and soft rough sets [5].

**Acknowledgements**

This work was supported by the National Natural Science Foundation of China under Grants 61070251, 61170270, and 61121061 and the Fundamental Research Funds for the Central Universities under Grant 2012RC0710.